\documentclass{article}
\usepackage{CJKutf8}

\usepackage{PRIMEarxiv}

\usepackage[utf8]{inputenc} 
\usepackage[T1]{fontenc}    
\usepackage{hyperref}       
\usepackage{url}            
\usepackage{booktabs}       
\usepackage{amsfonts}       
\usepackage{nicefrac}       
\usepackage{lipsum}
\usepackage{graphicx}
\usepackage{longtable}      
\usepackage{calc}           
\usepackage{float}          
\graphicspath{{media/}}     

\title{ScoreRAG: A Retrieval-Augmented Generation Framework with Consistency-Relevance Scoring and Structured Summarization for News Generation
}

\author{
  Pei-Yun Lin \\
  Department of Mathematical Sciences, National Chengchi University\\
  Taiwan\\
  \texttt{111751002@nccu.edu.tw}
  \and
  Yen-lung Tsai \\
  Department of Mathematical Sciences, National Chengchi University\\
  Taiwan\\
  \texttt{yenlung@nccu.edu.tw}
}

\begin{document}
\begin{CJK}{UTF8}{bsmi} 
\maketitle

\begin{abstract}
  This research introduces ScoreRAG, an approach to enhance the quality of automated news generation. Despite advancements in Natural Language Processing and large language models, current news generation methods often struggle with hallucinations, factual inconsistencies, and lack of domain-specific expertise when producing news articles. ScoreRAG addresses these challenges through a multi-stage framework combining retrieval-augmented generation, consistency relevance evaluation, and structured summarization. The system first retrieves relevant news documents from a vector database, maps them to complete news items, and assigns consistency relevance scores based on large language model evaluations. These documents are then reranked according to relevance, with low-quality items filtered out. The framework proceeds to generate graded summaries based on relevance scores, which guide the large language model in producing complete news articles following professional journalistic standards. Through this methodical approach, ScoreRAG aims to significantly improve the accuracy, coherence, informativeness, and professionalism of generated news articles while maintaining stability and consistency throughout the generation process. The code and demo are available at: \url{https://github.com/peiyun2260/ScoreRAG}
\end{abstract}


\section{Introduction}
Recent advancements in Natural Language Processing (NLP), particularly the development of transformer-based 
architectures \cite{1706.03762}, have significantly improved the capacity of language models to generate 
fluent and coherent text. This progress has opened the door for many real-world applications, including 
automatic news generation. However, despite their impressive performance, large language models (LLMs) 
still face key limitations when applied to tasks requiring factual precision and domain-specific knowledge.

In zero-shot or instruction-based text generation, LLMs frequently suffer from hallucinations, factual 
In zero-shot or instruction-based generation settings, LLMs frequently suffer from hallucinations, factual inconsistencies, and stylistic deviations from journalistic norms \cite{10.1145/3703155, 2401.11817, ji2023hallucination}. These limitations are particularly detrimental in the news domain, where even minor inaccuracies can undermine credibility. Existing decoding techniques, such as temperature scaling and top-$k$/top-$p$ sampling, provide limited control over factual alignment and structural consistency.

To address these issues, recent work has introduced techniques that incorporate external information and reasoning capabilities. Retrieval-Augmented Generation (RAG) enhances generation by grounding responses in retrieved evidence \cite{lewis2020rag}, while reranking mechanisms improve the relevance and quality of selected documents \cite{li2023re2g}. Furthermore, planning-based prompting methods such as Plan-and-Solve \cite{yao2023planandsolve} and self-consistency strategies \cite{wang2023selfconsistency} have shown promise in improving reasoning and stability.

Building on these insights, this paper proposes ScoreRAG, a novel framework that integrates retrieval-based grounding, consistency scoring, and structured summarization to enhance automated news generation. ScoreRAG consists of the following components:

\begin{itemize}
\item
  \textbf{RAG-based News Retrieval:} Retrieves relevant news documents
  from a database to ground the generation process.
\item
  \textbf{Mapping News from Database:} Links the retrieved information
  to structured news items to obtain complete articles.
\item
  \textbf{Consistency Scoring and Reranking:} Assigns consistency scores to 
  retrieved documents based on LLM evaluations, reranks them according to relevance, 
  and filters out low-quality items before summarization.
\item
  \textbf{Score-based Summarization Generation:} Creates graded
  summaries according to consistency scores to enhance informativeness
  and accuracy.
\item
  \textbf{Guided News Generation:} Guides the LLM with structured
  instructions to generate complete news articles based on the optimized
  summaries.
\end{itemize}

Through this multi-stage process, ScoreRAG ensures that generated news
articles are not only fluent and coherent but also factually accurate
and professionally structured.

\section{Methods}

This chapter introduces the proposed framework, ScoreRAG, designed to
enhance retrieval-augmented generation through self-consistency-based
reranking mechanism, and a score-guided summarization process to
generate more accurate and coherent outputs.

\begin{figure}[h]
  \centering
  \includegraphics[width=0.8\textwidth]{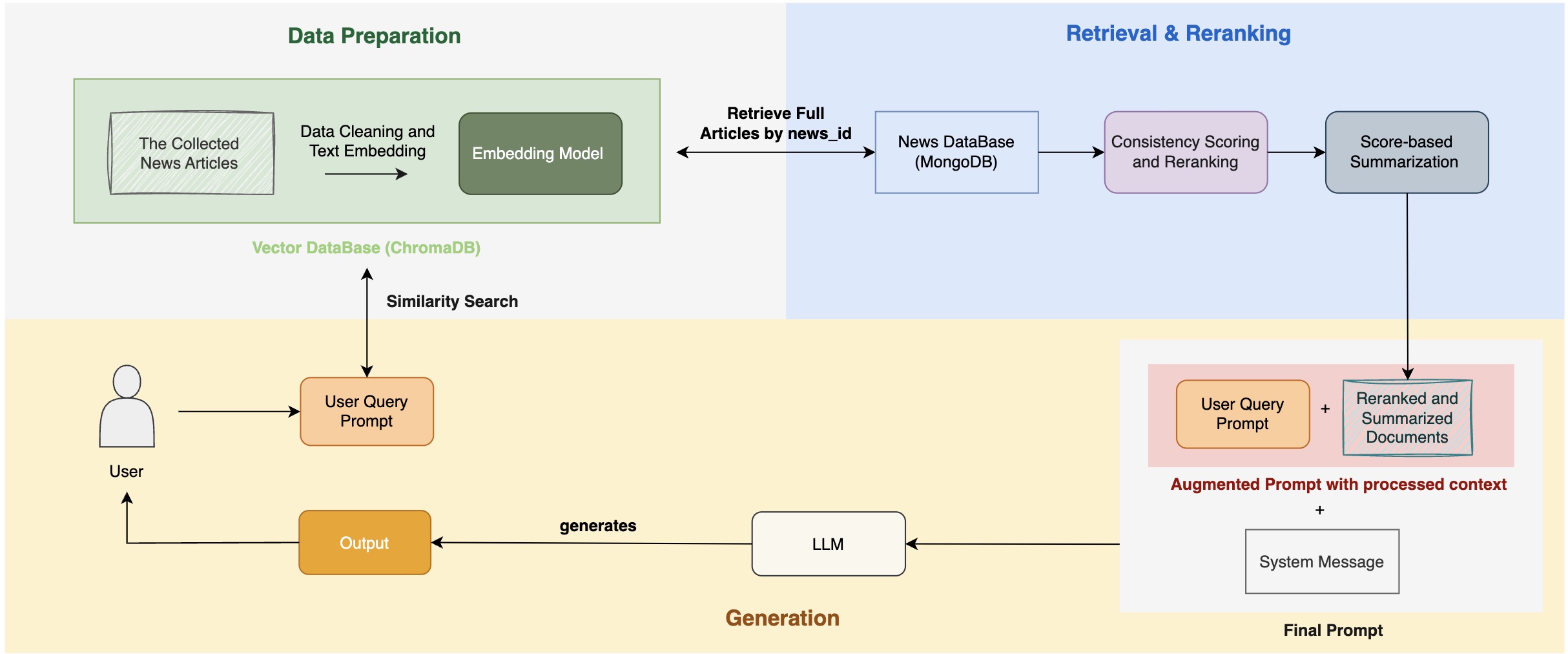}
  \caption{The overall system architecture of ScoreRAG}
  \label{fig:ScoreRAG}
\end{figure}

The overall system architecture of ScoreRAG is illustrated in figure \ref{fig:ScoreRAG}.
The green block on the left represents the data preparation stage, which
includes data cleaning and text embedding. The blue block on the right
corresponds to the retrieval and reranking stages, where the retrieved
documents are reranked and summarized into graded outputs. The final
stage, depicted by the yellow block at the bottom, augments the LLM
prompt with the processed context to generate the final output. The
subsequent sections detail each step of the system architecture.

\subsection{Data Preprocessing}\label{data-preprocessing}

\subsubsection{Data Cleaning}\label{data-cleaning}

\begin{figure}[h]
  \centering
  \includegraphics[width=0.6\textwidth]{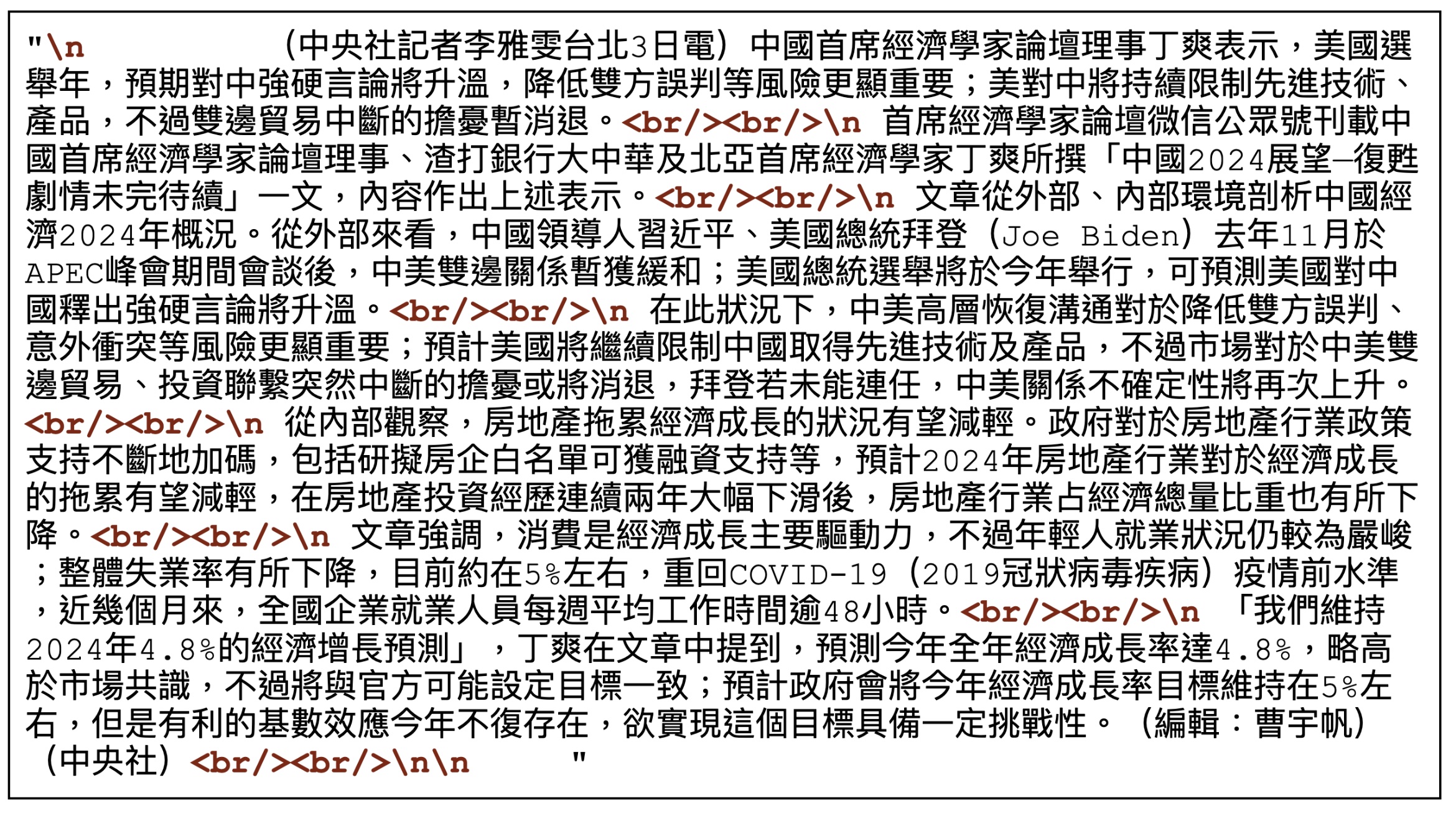}
  \caption{The collected news that has not been cleaned}
  \label{fig:collectedData}
\end{figure}

To enhance the quality of news articles collected through web crawlers---often cluttered with irrelevant content that can negatively impact the effectiveness of word embeddings, such as HTML tags or advertisements in article (Figure \ref{fig:collectedData})---preprocessing the data is essential. It needs to remove HTML tags, advertisements, and other irrelevant content to ensure that only the core news text is preserved for analysis.

In the final dataset, after data cleaning, each entry will consist of
the following fields: \emph{news\_id, published\_date, title, summary
and content.}

{\tiny
\begin{longtable}{@{}p{0.1cm} p{0.8cm} p{1cm} p{3.3cm} p{4.2cm} p{6cm}@{}}
  \caption{\label{tab:news_dataset}News dataset includes news\_id, date, title, summary, and content.} \\
  \toprule
  \textbf{No.} & \textbf{news\_id} & \textbf{date} & \textbf{news\_title} & \textbf{news\_summary} & \textbf{news\_content} \\
  \midrule
  \endfirsthead
  
  \toprule
  \textbf{No.} & \textbf{news\_id} & \textbf{date} & \textbf{news\_title} & \textbf{news\_summary} & \textbf{news\_content} \\
  \midrule
  \endhead
  
  \bottomrule
  \endlastfoot
  \tiny
  0 & 2384857 & 2024-02-01 & 美官員：與中方芬太尼會談有意義但尚須更多措施 & 美中官員在北京會晤，就阻止芬太尼化學品流入美國舉行磋商。白宮代表團團長表示，會談有實質性，但... & （德國之聲中文網）美國總統副助理、白宮國土安全事務副助理達斯卡爾（Jen Daskal）表示... \\
  
  1 & 2384858 & 2024-02-01 & 德語媒體：恆大的骨牌效應 & 周一的香港法庭對恆大集團發出的清盤令，對股市造成沖擊。而這一裁決將會對中國房地產市場、金融行... & （德國之聲中文網）《南德意志報》發表評論稱，恆大鬧劇終於告一段落，但投資者的不安全感卻在進一... \\
  
  2 & 2384859 & 2024-02-01 & 德國經銷商柏林開設比亞迪門市 & 德國經銷商 Sternauto 在柏林新開設專賣店，展示中國電動車巨頭比亞迪在德國銷售的五款車型... & （德國之聲中文網）Sternauto 在德國東部地區擁有比亞迪汽車獨家銷售權。隨著... \\
  
  3 & 2384860 & 2024-02-01 & 中國在秘魯建巨型港口 連接亞洲與南美 & 在秘魯漁港城市錢凱，一座由中國興建的巨型港口正在崛起，以期加快南美洲與中國之間的貿易。 & （德國之聲中文網）這座耗資 35 億美元的深水港預計將於今年底開始運作，為中國提供直接的門戶，獲... \\
  \end{longtable}}    

I collect a total of news data from January 2018 to June 2024, amounting
to approximately 30,000 articles per month, resulting in a total of
2,310,000 articles.

\subsubsection{Text Embeddings}\label{text-embeddings}

Text embeddings serve as fundamental components in information retrieval
systems and retrieval augmented language models. However, most models
are trained only on English, so I choose \emph{multilingual-e5-model} \cite{2402.05672}
for embedding model which is a model initialized from
\emph{xml-roberta-large} model \cite{1911.02116}, a transformer-base architecture, and
continually trained on a mixture of multilingual datasets. This model
consists of 24 layers, with an embedding size of 1024.

I use ChromaDB \cite{ChromaGitHub}, a vector database designed to store and retrieve
unstructured data as high-dimensional vectors, enabling fast similarity
searches. ChromaDB integrates seamlessly with LangChain, making it an
ideal choice for managing text embeddings.

Before storing the data, I need to convert the news data into a
Document, the required data type for ChromaDB. A Document is a
structured list containing the following components:

\begin{itemize}
\item
  \textbf{metadata} (optional): Descriptive information about the news
  data, which can be used for filtering during queries. In our
  application, I stored the published data, news title and news\_id.
\item
  \textbf{page\_content}: The main text content of the document. In our
  application, I stored news content.
\end{itemize}

This document structure ensures proper formatting for efficient storage
and retrieval in ChromaDB. After formatting, the example news above will
be converted as:

\begin{figure}[H]
	\begin{center}                                                       
	\includegraphics[width=0.7\textwidth]{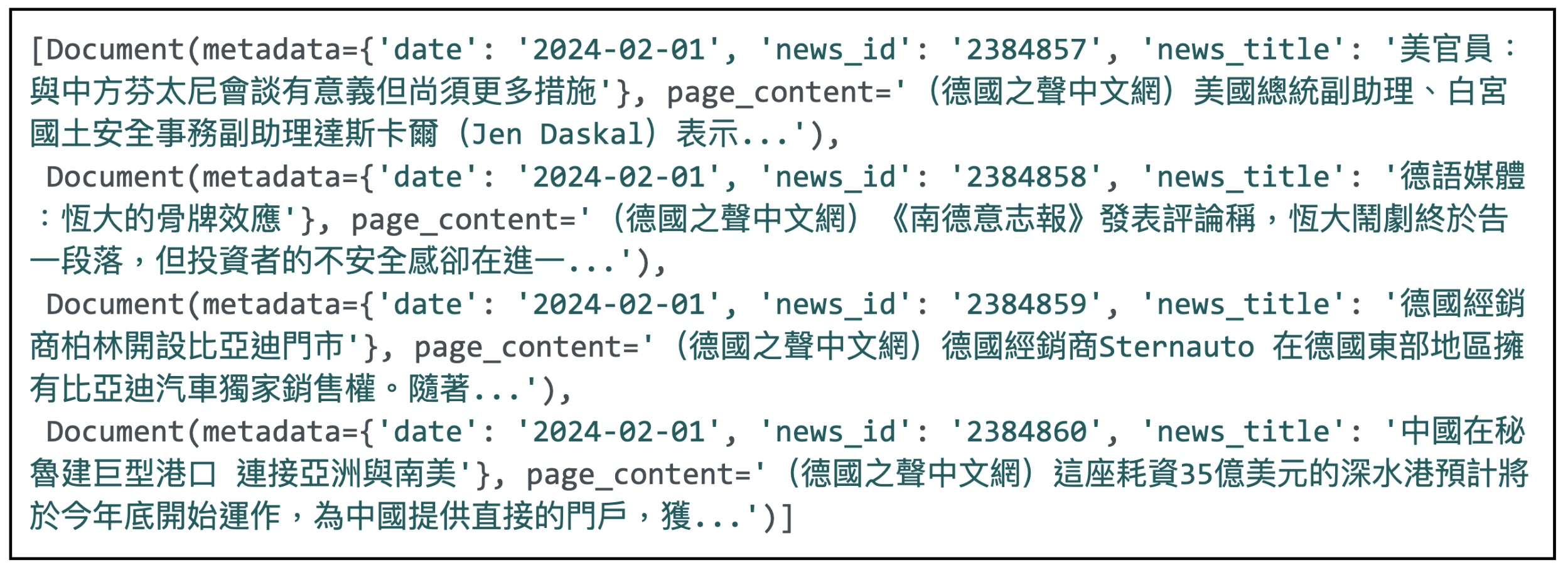}
	\caption{The Document data type}
	\label{fig:documentType}
	\end{center}
\end{figure}

Before applying word embedding, it is necessary to split the news
articles to meet the token limit required for embedding, ensuring that
the number of tokens stays within the model's capacity. I use the
\emph{RecursiveCharacterTextSplitter} class from LangChain's framework \cite{LangChainGitHub}
to perform the splitting. This splitter offers several configurable
parameters:

\begin{itemize}
\item
  \textbf{chunk\_size}: Controls the size of each chunk.
\item
  \textbf{chunk\_overlap}: Sets the number of overlapping characters
  between adjacent chunks to retain context and prevent loss of meaning
  during the split.
\item
  \textbf{length\_function}: Specifies the function used to calculate
  the length of each chunk.
\item
  \textbf{separators}: A list of delimiters that determines the order of
  splitting. The splitter tries each separator in sequence until it
  finds a suitable split point. The default separators are
  {[}"\textbackslash n\textbackslash n", "\textbackslash n", " ", ""{]},
  meaning it will first attempt to preserve paragraphs, followed by
  sentences, words, and finally individual characters.
\end{itemize}

After applying the text splitter, the input text will be divided into
multiple chunks, with overlapping content where needed to ensure smooth
continuity between adjacent chunks.

Finally, for every chunk of news data, I embedding it to
high-dimensional vectors through multilingual-e5-model embedding model
and store it on ChromaDB. I use the Euclidean distance to compute the
distances between vectors, and the formula is defined as:

\[Squared\ L2\ \ \ d = \sum_{i}^{n}{(x_{i} - y_{i})^{2}\ }\]

Where \(x_{i}\) and \(y_{i}\) are vectors that have a dimensionality of
1024, so \(n\) is 1024. Therefore, every time a user inputs the query, I
embed it and compute the distance with the all data in ChromaDB and then
return the top-k related data.

For embedding the processed text, the configuration allows setting the model name and computing device. The actual embedding is performed using the \emph{Chroma.from\_documents} function \cite{ChromaWebsite} with a batch size of 1000, enabling efficient large-scale processing. Detailed implementation can be found in \cite{scoreRAG2025}. \footnote{See \texttt{backend/app/services/embedding\_service.py}.} This improves computational efficiency, resource utilization and ensures the stability and performance of the model.

\subsection{System Architecture}\label{system-architecture}

\subsubsection{RAG-based News Retrieval}\label{rag-based-news-retrieval}

After data preprocessing, all the collected news articles are stored as
embedded vectors. When a query is provided, a similarity search is
performed between the input and the stored articles. For example, if the
input is “美中官員在北京會晤對臺灣總統大選影響”, ChromaDB calculates
the similarity between the input and all the articles. Additionally, I
can specify the number of related data points to retrieve. If I set k =
4, the system will return the top-4 most relevant data points. 

The retriever function used for similarity search is implemented in Python and available at GitHub repository\cite{scoreRAG2025}.\footnote{See \texttt{backend/app/db/chroma\_connector.py}.} It specifies the embedding function and the directory for the vector store. The retrieved documents follow the same structure as the embedded data and are returned as Document objects, each containing both metadata and content fields.

\subsubsection{Mapping News from Database}\label{mapping-news-from-database}

Although I retrieve relevant data from ChromaDB, each data point
represents only a chunk of a complete news article. Directly using these
chunks to augment LLM generation may result in incomplete or
insufficient responses due to missing context.

To address this issue, I utilize the \emph{news\_id} stored in the metadata
when indexing data in ChromaDB. This allows to map the retrieved chunks
back to the complete article stored in MongoDB \cite{MongoDB}, a NoSQL database that
uses a document-oriented storage model and stores data in a JSON-like
BSON format.

Since I have collected news articles from 2018 to 2024, I optimize
database search efficiency by storing each year's
articles in a separate collection. As a result, I maintain six
collections, each corresponding to a specific year from 2018 to 2024.

\begin{figure}[H]
	\begin{center}                                                       
	\includegraphics[width=0.7\textwidth]{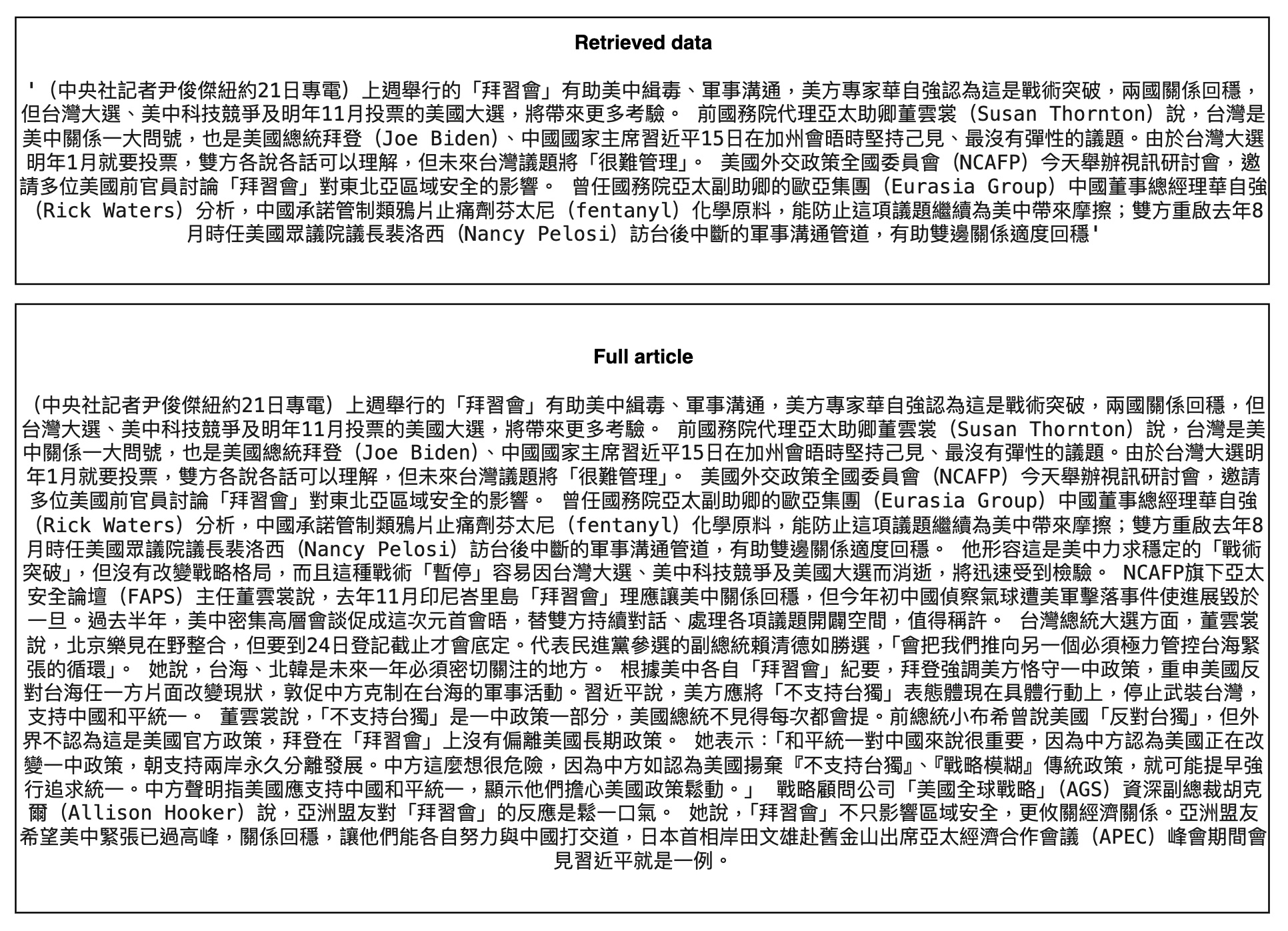}
	\caption{The retrieved data and the corresponding full article}
	\label{fig:fullArticles}
	\end{center}
\end{figure}

For example, as shown in Figure \ref{fig:fullArticles}, the upper section illustrates a retrieved chunk of text from an article, which may not provide sufficient context when presented alone. To enhance the completeness and relevance of the information, the full article corresponding to the chunk is retrieved, as shown in the lower section of Figure \ref{fig:fullArticles}. This additional context helps improve the accuracy of the information fed into the LLM.

\subsubsection{Consistency Scoring and Reranking}\label{Consistency-scoring-and-rerankin}

After retrieving relevant data, the initial ranking may not accurately reflect the semantic relevance of each document. Additionally, the amount of text may exceed the LLM's token limit, which can lead to context dilution. To mitigate this, I apply
a reranking step and generate a summary to replace the full article,
ensuring the LLM to focus on the most relevant content during
generation.

For reranking, I employ a self-consistency approach to ensure reliable relevance assessment. Since language models can produce varying outputs due to the inherent randomness in decoding strategies such as temperature sampling and top-$k$ sampling, individual evaluations may be inconsistent. To mitigate this variability, I implement a self-consistency mechanism \cite{wang2023selfconsistency} where the LLM evaluates the consistency score between the input query and each retrieved document three times, then computes the average score to obtain a more stable and reliable assessment. This approach leverages multiple sampling paths to converge on a more consistent evaluation result.

The LLM assigns a consistency score based on the date,
title, and summary of each article, following the criteria below:

\begin{itemize}
\item
  \textbf{90-100}: Highly relevant --- directly discusses the query
  topic.
\item
  \textbf{70-89}: Strongly relevant --- closely related to the query
  topic but may slightly deviate.
\item
  \textbf{50-69}: Somewhat relevant --- partially related but not the
  main focus.
\item
  \textbf{0-49}: Not relevant --- unrelated or minimally related to the
  query topic.
\end{itemize}

To avoid retrieving unrelated article, I set a threshold to filter out
any article with a score below 20. This approach ensures that only the most relevant content is included in the context, addressing both the issues of context dilution and model inconsistency. The complete function code is provided in \cite{scoreRAG2025}. \footnote{See \texttt{backend/app/services/CoT\_service.py}.}

\subsubsection{Score-based Summarization Generation}\label{score-based-summarization-generation}

Next, to generate a summary based on the consistency score, I design a
graded news summarization approach. The higher the score, the longer and
more detailed the summary, ensuring it captures the essential facts of
the article while maintaining relevance. The summarization criteria are
as follows:

\begin{itemize}
\item
  \textbf{Score above 70}: The summary must include core facts, key
  data, main quotes, background information, and impact assessment.
\item
  \textbf{Score between 50 and 70}: The summary must include core facts,
  key data, main quotes, and brief background information.
\item
  \textbf{Score between 30 and 50}: Only core facts and the most
  important data are required.
\item
  \textbf{Score between 20 and 30}: Include only the most essential core
  facts.
\end{itemize}

Additionally, I provide specific instructions to the LLM to ensure the
summary adheres to formal news standards, as outlined below:

\begin{itemize}
\item
  Preserve key facts, data, and important quotes (if applicable) from
  the original news article.
\item
  Maintain objectivity and avoid personal opinions.
\item
  Retain the original time and location information.
\item
  If multiple perspectives are presented, ensure the summary remains
  balanced.
\end{itemize}

For the reranking step and graded summary generation, I utilize the
LLaMA 3.1 8B model. To run the model, I leverage Ollama, which is
integrated with LangChain---a framework designed for building
applications powered by LLMs. This setup allows seamless deployment and
execution of the LLM directly on local machines. The complete function
code is provided in \cite{scoreRAG2025}. \footnote{See \texttt{backend/app/services/summary\_service.py}.}

\subsubsection{Guided News Generation}\label{guided-news-generation}

Finally, after retrieving relevant data and processing it through the
planning strategy, I feed the final articles into the LLM to augment
generation. To ensure clear referencing, I define a structured reference
format. In addition to the graded summary, each article includes the
date, article title, and its consistency score. This helps the LLM
accurately reference the sources during generation.

To ensure the LLM generates professional articles with accurate
referencing and allows users to clearly identify the source of each
paragraph, I provide the following specific instructions to the LLM:

\begin{itemize}
\item
  \textbf{Citation Format}: Use the format '(Reference
  X)', where X represents the reference number.
\item
  \textbf{Fact-Based Content}: All key information and data must be
  directly based on the references or the original input.
\item
  \textbf{Professional Language}: Write in a professional news reporting
  tone in Traditional Chinese. Avoid casual language or overly academic
  expressions.
\item
  \textbf{Time Stamping}: Clearly specify time points (e.g., "January
  2024"). Avoid vague or relative terms such as "last month".
\end{itemize}

The complete function code is provided in \cite{scoreRAG2025}. \footnote{See \texttt{backend/app/services/generation\_service.py}.}

To support the presentation of ScoreRAG, I developed a simple frontend interface as a demonstration platform. This interface allows users to input a news topic and the resulting article is displayed alongside clearly formatted references, which ensures that users can trace the source of each piece of information,  enhancing both transparency and traceability of the generated content. The interface code and example outputs are available in our GitHub repository\cite{scoreRAG2025}.\footnote{Frontend interface located at \texttt{frontend/src} and sample data in \texttt{example\_output/}.}

\section{Experiments}

\subsection{Experiment Setup}\label{experiment-setup}

\subsubsection{Compared Methods}\label{compared-methods}

To evaluate the effectiveness of the proposed ScoreRAG approach for the
news generation task, I compare it with a Zero-shot baseline, where the
LLM generates responses directly without any RAG-based augmentation.

\subsubsection{Evaluation Strategy}\label{evaluation-strategy}

To evaluate the quality of the generated article, I invited experts to
design a four-criteria evaluation framework, each criterion is rated on a 1-5 point
scale.

\begin{itemize}
\item
  \textbf{Coherence (1-5 points)}

  \begin{itemize}
  \item
    5 points: The report is complete and well-organized, covering all
    5W1H elements (What, Who, When, Where, Why, How). It presents a
    clear timeline, structured layers of information, and smooth,
    logical transitions.
  \item
    3 points: Two 5W1H elements are missing (e.g., key people, time, or
    place), or paragraph transitions are somewhat disjointed.
  \item
    1 point: The structure is confusing, with 4--5 missing 5W1H
    elements, inconsistent content, or a flawed timeline.
  \end{itemize}
\item
  \textbf{Professionalism (1-5 points)}

  \begin{itemize}
  \item
    5 points: The tone is objective and fully aligns with formal
    journalistic standards.
  \item
    3 points: The language occasionally feels colloquial or lacks a
    consistent professional tone.
  \item
    1 point: The article is unprofessional in tone or includes subjective
    bias.
\end{itemize}
\item
  \textbf{Informativeness (1-5 points)}

  \begin{itemize}
  \item
    5 points: Rich in detail, including background information, causes,
    effects, and key figures.
  \item
    3 points: Provides basic context but lacks key details or depth.
  \item
    1 point: Limited to surface-level descriptions with minimal
    information.
  \end{itemize}
\item
  \textbf{Accuracy (1-5 points)}

  \begin{itemize}
  \item
    5 points: All content is factually correct, with no hallucinations
    or incorrect assumptions.
  \item
    3 points: Minor factual errors or partial misunderstandings.
  \item
    1 point: Contains obvious inaccuracies or severe hallucinations.
  \end{itemize}
\end{itemize}

Since the ScoreRAG approach focuses on addressing hallucination problem
in LLM and enhancing background knowledge during news article
generation, accuracy and informativeness are prioritized. In contrast,
professionalism in writing is expected to show minimal variation across
methods.

Therefore, to achieve a reasonable and comprehensive evaluation of the
generated news articles, the final score is computed by aggregating the
individual dimension scores with predefined weights. The scores for
coherence, accuracy, professionalism, and informativeness are weighted
at 0.2, 0.35, 0.1, and 0.35, respectively.

To ensure both scalability and reliability in evaluating the quality of
generated news articles, I adopted a dual evaluation strategy combining
automated LLM-based scoring with manual expert assessment.

LLM based evaluation enables efficient, large-scale scoring across
multiple criteria, making it well-suited for assessing all 50 generated
articles. In contrast, expert evaluation offers deep contextual
understanding and news value judgment that automated methods may
overlook. To incorporate this human perspective, I engaged two
professional journalists, each independently reviewing a subset of 10
articles. This combined approach ensures the evaluation is both
comprehensive and credible.

\subsection{Results}\label{results}

\subsubsection{LLM Evaluation}\label{llm-evaluation}

Table~\ref{tab:llm_avg_scores} presents the average total scores of the ScoreRAG approach and the Zero-shot baseline, evaluated by LLM. The results show that ScoreRAG achieves a higher average score (4.64) compared to Zero-shot (4.34), indicating a consistent overall advantage. This suggests that the ScoreRAG approach improves the quality of generated content across various news topics.

\begin{table}[ht]
  \centering
  \caption{Average total scores of LLM evaluation}
  \label{tab:llm_avg_scores}
  \begin{tabular}{lc}
  \toprule
  Model       & Average Score \\
  \midrule
  ScoreRAG    & 4.64 \\
  Zero-shot   & 4.34 \\
  \bottomrule
  \end{tabular}
  \end{table}

\begin{figure}[H]
  \centering
  \begin{minipage}{0.48\textwidth}
    \centering
    \includegraphics[width=0.95\textwidth]{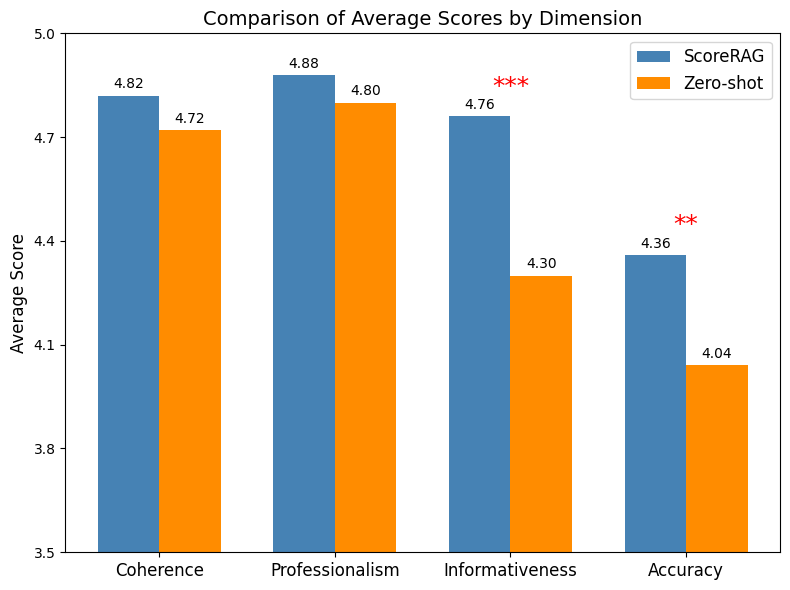}
    \caption{Average LLM evaluation scores across four dimensions}
    \label{fig:avg_llm}
  \end{minipage}
  \hfill
  \begin{minipage}{0.48\textwidth}
    \centering
    \includegraphics[width=0.95\textwidth]{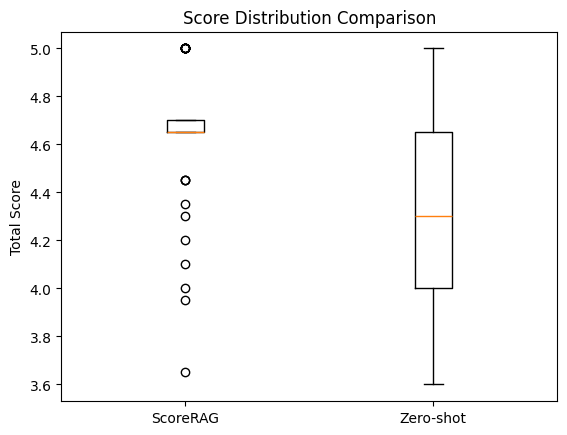}
    \caption{Boxplot of weighted total scores for LLM evaluation}
    \label{fig:box_llm}
  \end{minipage}
\end{figure}

Figure \ref{fig:avg_llm} presents the average scores across four evaluation dimensions.
The ScoreRAG method consistently outperforms the Zero-shot baseline
across all dimensions. Significant improvements are observed in
informativeness (p \textless{} 0.001) and accuracy (p \textless{} 0.01),
suggesting that the ScoreRAG method effectively enhances the factuality
correctness and richness of generated news articles.

Moderate gains are also noted in coherence and professionalism,
reflecting overall improvements in the linguistic quality and stylistic
appropriateness of the outputs. These results demonstrate that the
ScoreRAG framework improves both the content quality and the writing
style of generated news articles.

As shown in figure \ref{fig:box_llm}, the ScoreRAG produces significantly more
consistent results compared to the Zero-shot baseline. The boxplot
reveals that ScoreRAG scores are tightly clustered around a higher
median, indicating stable and reliable generation quality.

In contrast, the Zero-shot baseline exhibits a much wider distribution,
suggesting greater variability in generation performance. These results
demonstrate that the ScoreRAG method not only improves the average
quality but also enhances the consistency of news generation.

\subsubsection{Expert Evaluation}\label{expert-evaluation}

Table~\ref{tab:expert_avg_scores} presents the average total scores from expert evaluation for both the ScoreRAG approach and the Zero-shot baseline. ScoreRAG achieved a notably higher average score (3.83) compared to Zero-shot (3.08), indicating its consistent effectiveness and superiority in human-assessed quality across the evaluated news articles.

\begin{table}[ht]
  \centering
  \caption{Average total scores of expert evaluation}
  \label{tab:expert_avg_scores}
  \begin{tabular}{lc}
  \toprule
  Model       & Average Score \\
  \midrule
  ScoreRAG    & 3.83 \\
  Zero-shot   & 3.08 \\
  \bottomrule
  \end{tabular}
  \end{table}

\begin{figure}[H]
  \centering
  \begin{minipage}{0.48\textwidth}
    \centering
    \includegraphics[width=0.95\textwidth]{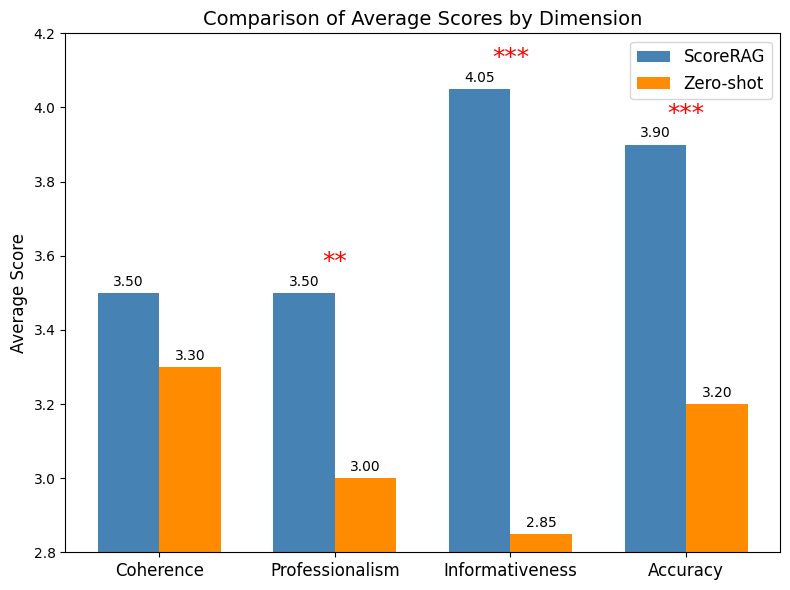}
    \caption{Average expert evaluation scores across four dimensions}
    \label{fig:avg_expert}
  \end{minipage}
  \hfill
  \begin{minipage}{0.48\textwidth}
    \centering
    \includegraphics[width=0.95\textwidth]{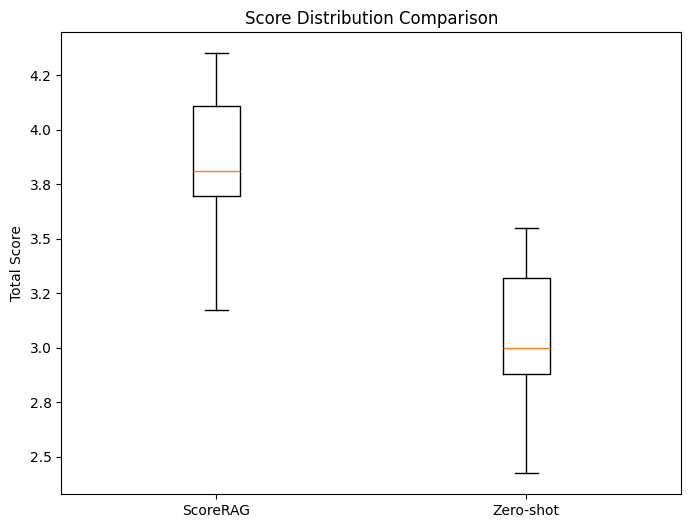}
    \caption{Boxplot of weighted total scores for expert evaluation}
    \label{fig:box_expert}
  \end{minipage}
\end{figure}

Figure \ref{fig:avg_expert} presents the average scores across four evaluation dimensions.
The ScoreRAG method consistently outperforms the Zero-shot baseline
across all dimensions. Significant improvements are observed in
professionalism (p \textless{} 0.01), informativeness (p \textless{}
0.001), and accuracy (p \textless{} 0.001), indicating that the ScoreRAG
approach produces more professional, informative, and accurate news
articles.

Figure \ref{fig:box_expert} presents the boxplot comparison of total scores between the
ScoreRAG and Zero-shot methods. The ScoreRAG method exhibits a higher
median score compared to the Zero-shot baseline. Furthermore, the two
distributions show no overlap, indicating a clear performance advantage
of interquartile.

\subsection{Analysis}\label{analysis}

In conclusion, both LLM evaluation and expert evaluation show that the
ScoreRAG approach outperforms the Zero-shot baseline across all
dimensions, particularly in accuracy and informativeness, which are
critical metrics for news generation.

Overall, LLM evaluation scores are generally higher, likely because LLMs
lack the expected standards and reference experience that human experts
possess. As a result, improvements in professionalism are more
noticeable in expert evaluation, whereas LLMs tend to assign high scores
as long as there are no obvious errors.

Regarding the moderate improvement in coherence, it can be attributed to
the language model's inherent strength in maintaining linguistic fluency
and stylistic consistency. Therefore, the difference between LLM
evaluation and expert evaluation for coherence is less pronounced,
though the ScoreRAG approach still demonstrates better performance.

Additionally, the ScoreRAG approach shows greater stability, which is
particularly important for news generation compared to other text
generation tasks, as news writing demands a high level of
professionalism and consistency.

\section{Conclusion}\label{conclusion}

This research proposed the ScoreRAG, a retrieval-augmented news
generation method enhanced with consistency scoring and reranking and
score-based summarization. Experiments results show that both LLM
evaluation and expert evaluation demonstrate the ScoreRAG significantly
outperforms the Zero-shot baseline across key metrics, particularly in
accuracy and informativeness.

Although improvement in coherence and professionalism is more moderate
in LLM evaluation, expert evaluation reveals a notable enhancement in
professionalism. Given that the expert evaluators are practicing
journalists, this finding indicates that the ScoreRAG is more capable of
generating news articles that meet real-world professional standards,
thereby enhancing its practical applicability in actual news writing
scenarios.

Overall, the ScoreRAG approach not only improves factual correctness but
also provides greater stability, which is crucial for news generating
tasks requiring high consistency and domain-specific standards.

\section{Future Work}\label{future-work}

During the experimental phase, it was observed that LLM scores and expert scores were sometimes misaligned. In some cases, expert ratings were higher, but LLM scores did not necessarily reflect this. Additionally, there were discrepancies in expert ratings, with different evaluators potentially leading to inconsistent score distributions due to subjective factors. In the future, I plan to invite more expert evaluators, such as journalists from varied domains to establish a robust benchmark for training and fine-tuning the LLM scoring mechanism, to enhance the LLM's alignment with professional standards, ensuring that its evaluations better reflect real-world journalistic expectations.

Furthermore, due to the increasing prevalence of fake news, many unverified stories have been widely circulated online, making it challenging to distinguish truth from misinformation. In the future, I aim to enhance the ScoreRAG framework by incorporating it as a fact-checking tool to assist in fake news detection, helping to address the ongoing issue of misinformation.

Beyond fake news detection, ScoreRAG's potential extends to other domains. For instance, the framework could be adapted for educational content generation, producing factually accurate and professionally written materials for academic or training purposes. Similarly, its application in financial news generation could ensure high standards of precision and professionalism, catering to industries with stringent requirements. Future research will explore these cross-disciplinary applications, validating ScoreRAG's versatility in diverse professional settings.

In the long term, ScoreRAG has the potential to transform the news industry by enhancing the credibility and professionalism of AI-generated articles. By aligning with journalistic ethics and standards, the framework could contribute to rebuilding public trust in media, addressing the challenges of information overload and misinformation. Collaborations with news organizations will be pursued to deploy ScoreRAG in real-world industries, gathering practical feedback to refine its performance. Additionally, open-sourcing the ScoreRAG framework is planned to encourage global research collaboration, fostering continuous improvements and broader adoption.

Through these efforts, ScoreRAG aims to not only advance the state-of-the-art in news generation but also make a meaningful societal impact by promoting accurate, professional, and ethical AI-driven journalism.

\bibliographystyle{unsrt}  
\bibliography{references}  

\end{CJK} 

\end{document}